\documentclass[final,3p,times]{elsarticle}

\usepackage{amssymb}
\usepackage{amsmath}
\usepackage{algorithm}
\usepackage{algpseudocode}
\usepackage{microtype}
\usepackage{booktabs}
\usepackage{lineno}
\usepackage{graphicx}
\usepackage{hyperref}

\journal{TBD}

\begin{document}

\begin{frontmatter}

\title{Interpretable long-term traffic modelling on national road networks using theory-informed deep learning}

\author[1]{Yue Li\corref{cor1}\fnref{eqcontrib}}
\ead{yl901@cam.ac.uk}
\author[1]{Shujuan Chen\fnref{eqcontrib}}
\author[2]{Akihiro Shimoda}
\author[1]{Ying Jin}

\cortext[cor1]{Corresponding author.}
\fntext[eqcontrib]{These authors contributed equally to this work.}

\affiliation[1]{organization={Martin Centre for Architectural and Urban Studies, University of Cambridge},
            addressline={1--5 Scroope Terrace},
            city={Cambridge},
            postcode={CB2 1PX},
            country={United Kingdom}}
\affiliation[2]{organization={Independent Researcher},
            city={Cambridge},
            country={United Kingdom}}

\begin{abstract}
Long-term traffic modelling is fundamental to transport planning, but existing approaches often trade off interpretability, transferability, and predictive accuracy. Classical travel demand models provide behavioural structure but rely on strong assumptions and extensive calibration, whereas generic deep learning models capture complex patterns but often lack theoretical grounding and spatial transferability, limiting their usefulness for long-term planning applications. We propose DeepDemand, a theory-informed deep learning framework that embeds key components of travel demand theory to predict long-term highway traffic volumes using external socioeconomic features and road-network structure. The framework integrates a competitive two-source Dijkstra procedure for local origin–destination (OD) region extraction and OD pair screening with a differentiable architecture modelling OD interactions and travel-time deterrence. The model is evaluated using eight years (2017–2024) of observations on the UK strategic road network, covering 5,088 highway segments. Under random cross-validation, DeepDemand achieves an $R^2$ of 0.718 and an MAE of 7,406 vehicles, outperforming linear, ridge, random forest, and gravity-style baselines. Performance remains strong under spatial cross-validation ($R^2 = 0.665$), indicating good geographic transferability. Interpretability analysis reveals a stable nonlinear travel-time deterrence pattern, key socioeconomic drivers of demand, and polycentric OD interaction structures aligned with major employment centres and transport hubs. These results highlight the value of integrating transport theory with deep learning for interpretable highway traffic modelling and practical planning applications.

\begin{abstract}

\end{abstract}
\end{abstract}


\end{frontmatter}

\section{Introduction}
\label{sec:intro}

Traffic modelling is fundamental to transport planning and policy. Road transport accounts for the majority of passenger travel, and vehicle ownership continues to rise globally, particularly in rapidly urbanising regions \cite{govuk_transport_2022,oecd_transport_2023,vehicle_ownership}. As a result, accurate estimates of traffic volumes are essential for infrastructure investment and maintenance planning, congestion mitigation, air-quality and carbon emission regulation, and the assessment of environmental and public health impacts \cite{IPCC2022_Mitigation, airquality, emission, trafficvolumedecisionmaking, chen_parta}. Robust traffic prediction therefore forms a critical foundation for evidence-based decision-making in transport systems.

In recent years, much of the research in this area has focused on short-term traffic forecasting, where models predict traffic states minutes or hours ahead using historical sensor data \cite{DCRNN,STGCN,STGAT,Pan2019,Pan2022}. These approaches are highly effective for real-time monitoring and operational control. However, they are less suited to long-term planning tasks. Planning problems often involve future population change, land-use development, or infrastructure modification, and must frequently be addressed in locations where historical traffic counts are unavailable or no longer representative \cite{whytraditionalforecastingfail, shorttermtrafficforecasting, TFflowReview}. In such settings, trustworthy predictions must rely on external socioeconomic drivers and network structure, rather than on extrapolating recent observations, and should remain interpretable and transferable across space \cite{callforimprovement}.

For long-term travel demand forecasting, the standard approach is the four-step travel demand model (TDM) \cite{4step}, which decomposes travel behaviour into trip generation and attraction, trip distribution, mode choice, and traffic assignment. In principle, the TDM provides a clear mapping from zonal socioeconomic inputs to an origin-destination (OD) matrix and then to link-level flows. Its modular structure and behavioural grounding have made it a cornerstone of planning practice for decades. In practice, however, TDM implementations often rely on restrictive functional forms, coarse spatial zoning, and fixed impedance assumptions. The sequential structure of the TDM also treats its steps as largely independent, even though destination choice, mode choice, and route choice are closely interrelated in reality \cite{4stepshortcomings}. As a result, TDM-based workflows typically require repeated calibration of the OD matrix and assignment parameters, which is labour-intensive and increasingly difficult at large spatial scales or under novel planning scenarios \cite{4stepcalibration}.

Motivated by these limitations, recent studies have explored machine learning as a surrogate for individual steps of the TDM. Deep learning–based gravity models have been proposed to replace traditional trip distribution formulations and directly predict OD flows from zonal attributes \cite{DeepGravity}. Other work has focused on zone-to-zone travel demand forecasting using data-driven models \cite{zonetozone}, mode choice prediction using machine learning classifiers \cite{modechoicereplacement}, or approximating traffic assignment and equilibrium behaviour using graph neural networks \cite{assignmentreplacementgat}. These studies demonstrate that machine learning can improve flexibility and predictive performance for specific components of the demand modelling pipeline. However, most existing approaches replace only a single step of the TDM, are often difficult to interpret or explain in behavioural terms, and rarely produce link-level traffic volumes in a fully integrated and scalable manner.

These observations highlight a clear gap between existing modelling approaches. On the one hand, TDM-based models offer interpretability and theoretical structure but struggle with flexibility, scalability, and calibration effort. On the other hand, data-driven models can capture complex nonlinear relationships but often lack explicit network reasoning, behavioural transparency, or full coverage of the demand-to-flow pipeline. There remains a need for a modelling framework that combines the structured logic of the TDM with modern machine learning, operates directly at the level of highway segments, and is suitable for long-term planning applications.

In this study, we propose DeepDemand, a data-driven reimagination of the TDM for highway traffic volume prediction. The model mirrors the core steps of the TDM within a single differentiable framework trained directly against observed link volumes. Our aims are to (1) develop a theory-informed and interpretable model that predicts highway traffic volumes using static socioeconomic drivers and network structure, (2) evaluate its predictive performance and spatial transferability at national scale, and (3) analyse the learned model components using explainable AI techniques to extract behavioural and policy-relevant insights.

\section{Methodology}\label{sec:methodology}

\begin{figure}[htbp]
    \centering
    \makebox[\textwidth][c]{%
        \includegraphics[width=0.95\textwidth]{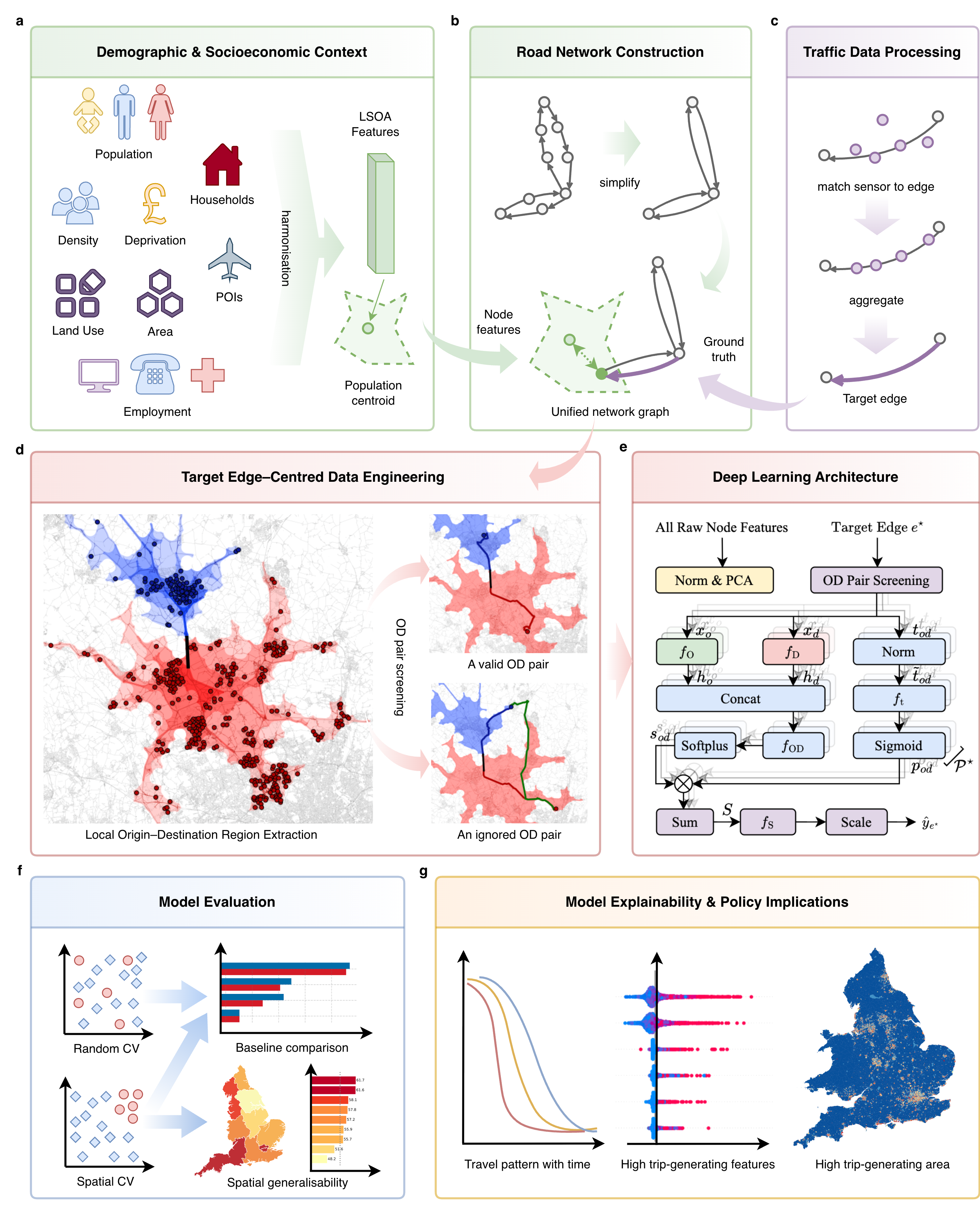}
    }
    \caption{\textbf{Framework overview.}
\textbf{a}, Demographic and socioeconomic features are harmonised to small-area (LSOA) units. 
\textbf{b}, A simplified road network is constructed with node features and ground truth attached. 
\textbf{c}, Traffic sensor records are matched and aggregated at edge level as ground truth. 
\textbf{d}, For each target edge, a local OD region is extracted and candidate OD pairs are screened. 
\textbf{e}, A deep learning model learns to predict traffic volume on the target edge. 
\textbf{f}, Model performance is evaluated using random and spatial cross-validation and compared with baseline models. 
\textbf{g}, Interpretability analyses reveal travel patterns, influential features, and spatial demand distributions.}
    \label{fig:framework}
\end{figure}

To address the research aims, we developed a five-stage framework, illustrated in Figure~\ref{fig:framework}. The framework comprises data collection and preprocessing, local OD region extraction and OD pair screening, deep learning model training, model evaluation, and model explainability with policy insight extraction. Multiple technical pathways were explored and evaluated during model development, and several methodological challenges were identified before converging on the final framework. A detailed discussion of the key considerations, limitations, and alternative approaches examined during this process is provided in Supplementary Appendix~S1.

\subsection{Data Collection and Preprocessing}\label{subsec:data}

The data used in this study are divided into two categories: input data and ground-truth data. The input data include a full UK driving network graph and a suite of demographic and socioeconomic features aggregated for small statistical areas in England and Wales. The ground-truth data consist of observed highway traffic volumes derived from the National Highways Traffic Information System (TRIS).

The UK driving network graph serves as the backbone of the model. It contains a set of nodes and edges with attributes such as geographic coordinates, topology, and feature labels. For small-area features, we adopt the Lower Layer Super Output Area (LSOA), which is a census-based spatial unit designed to contain roughly 1,000–1,500 residents. As of 2021, there are 35,672 LSOAs across England and Wales.

\begin{figure}[htbp]
    \centering
    \makebox[\textwidth][c]{
        \includegraphics[width=1\textwidth]{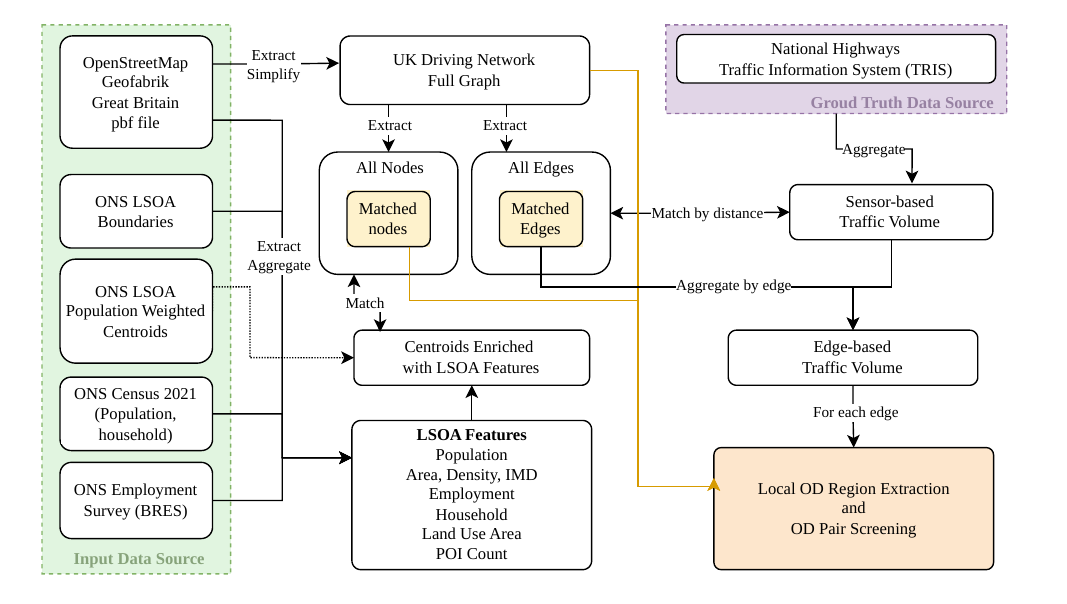}
    }
    \caption{\textbf{Data collection and preprocessing workflow.}}
    \label{fig:data}
\end{figure}

As illustrated in Figure~\ref{fig:data}, the data preparation process proceeds as follows. A full UK driving network is extracted and simplified into a topologically connected directed graph comprising nodes and edges. LSOA-level features are aggregated and assigned to the nearest graph nodes based on the shortest distance to population-weighted LSOA centroids. Ground-truth traffic volumes from TRIS sensors are spatially matched to and aggregated by graph edges. At the end of this process, we obtain a unified large graph dataset with a subset of nodes enriched with LSOA features and a subset of edges annotated with ground-truth traffic volumes. This enriched driving network serves as the foundation for subsequent local OD region extraction and OD pair screening.

\subsubsection{Driving Network}\label{subsubsec:network}

The national driving network used in this study was derived from the OpenStreetMap (OSM) Great Britain dataset obtained from Geofabrik (snapshot: 2025-01-01) \cite{geofabrik}. The raw OSM archive in PBF format was first filtered using \texttt{osmium} to retain only elements carrying the \texttt{highway} tag. The filtered dataset was then processed using \texttt{pyrosm} to extract the drivable road network, retaining road types accessible to motorised traffic.

The extracted road network was represented as node and edge GeoDataFrames and converted into a directed NetworkX graph using \texttt{osmnx}. During this conversion, non-essential attributes were removed to reduce memory footprint and computational overhead, while preserving key geometric and semantic attributes required for routing and modelling. These retained attributes include road geometry, road class (OSM \texttt{highway} tag), number of lanes (where available), posted speed limits (\texttt{maxspeed}), and edge length.

To obtain a topologically consistent and computationally manageable representation at national scale, the graph was simplified using \texttt{ox.simplify\_graph(strict=True)}. This procedure merges consecutive edges that share identical attributes and removes intermediate nodes that do not represent true intersections, while preserving network connectivity and directionality. The resulting network is represented as a graph $\mathcal{G} = (\mathcal{N},\,\mathcal{E})$, where the node set $\mathcal{N}=\{n\}$ corresponds to road junctions and endpoints, and the edge set $\mathcal{E}=\{e\}$ represents directed road segments. After simplification, the graph contains approximately 6.3 million nodes and 7.2 million directed edges.

\begin{figure}[htbp]
    \centering
    \makebox[\textwidth][c]{%
        \includegraphics[width=1.05\textwidth]{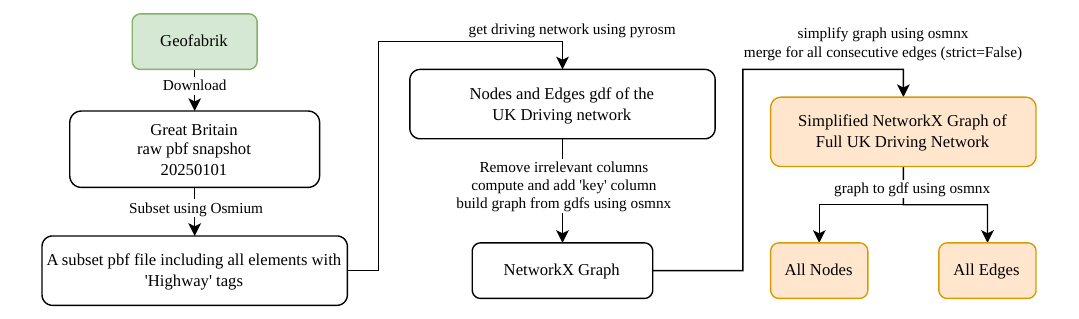}
    }
    \caption{\textbf{Construction and preprocessing of the UK national driving network.} The raw OSM Great Britain extract is filtered to retain only drivable road segments, converted into a directed graph with key geometric and semantic attributes, simplified to a topologically consistent national-scale network, and exported for downstream spatial joins and modelling.}
    \label{fig:uk_network}
\end{figure}

Both nodes and edges were exported as GeoJSON files to support spatial joins and feature mapping in subsequent processing steps. In addition, the full graph object was preserved in \texttt{.gpickle} format to retain structural metadata and enable efficient reuse across analyses. The overall workflow for constructing and preprocessing the UK national driving network is summarised in Figure~\ref{fig:uk_network}.

Each directed edge $(u,v)$ was then assigned a travel-time weight $t_{uv}$ based on its geometric length and an assumed travel speed. When an explicit \texttt{maxspeed} attribute was present in the OSM data, this value was used directly as the travel speed. When no valid speed limit was available, a fallback speed was assigned according to the OSM road class (\texttt{highway} tag), following a predefined mapping reflecting typical UK driving conditions. The fallback speeds (in miles per hour) used in this study were as follows: motorway 70, motorway\_link 60; trunk 60, trunk\_link 50; primary 45, primary\_link 40; secondary 35, secondary\_link 30; tertiary 25, tertiary\_link 20; residential 15, unclassified 15, service 15; and all other road types 30. Edge travel times were computed as the ratio of edge length to assigned travel speed, with appropriate unit conversion applied. The resulting travel times $t_{uv}$ serve as edge weights for all subsequent network analyses, including shortest-path computations, local origin--destination region extraction, and OD pair screening.

\subsubsection{LSOA Features}\label{subsubsec:lsoa}

As summarised in Table~\ref{tab:lsoa_features}, a set of socioeconomic, demographic, and environmental features was compiled at the LSOA level to characterise the spatial context of each small area. These features were derived primarily from the Office for National Statistics (ONS) and OSM and represent the main determinants of trip generation, attraction, and distribution in the TDM~\cite{4step}. They are also widely recognised in the transport and urban economics literature as core drivers of travel demand, systematically influencing trip generation, spatial interaction patterns, and traffic flows across networks \cite{traveldemand3D, Duranton2011, Wilson1970Entropy, wegener2004}.

\begin{table}[htbp]
\centering
\small
\caption{\textbf{Summary of LSOA-level features used for node feature enrichment.}}
\label{tab:lsoa_features}
\hspace*{-15mm}
\begin{tabular}{p{3cm} p{6.5cm} p{1.6cm} p{3cm} p{3cm}}
\toprule
\textbf{Feature group} & \textbf{Stratification and subcategories} & \textbf{Total features} & \textbf{Relevance to TDM} & \textbf{Data source} \\
\midrule
Area and population density &
Land area (1) and population density (1) &
2 &
Trip generation; Trip distribution &
ONS \cite{areapopden} \\
Population &
Total (1); stratified by sex (2) and age group (5: 0--15, 16--24, 25--49, 50--64, $\geq$65); stratified by sex--age combinations (10) &
18 &
Trip generation &
ONS \cite{population} \\
Employment &
Total (1); stratified by work type (2: full-time, part-time) and industrial sector (18); stratified by work type--sector combinations (36) &
57 &
Trip attraction &
ONS \cite{employment} \\
Household and car ownership &
Total (1); stratified by car/van availability (2: no cars or vans, one or more cars or vans) and household composition (10); stratified by car--household combinations (20) &
33 &
Trip generation; Modal split &
ONS \cite{households} \\
Income and employment deprivation &
2019 domain raw scores for income deprivation and employment deprivation &
2 &
Trip generation; Trip distribution &
IMD 2019 \cite{imd2019} \\
Land-use area &
Area of land-use polygons by class: residential, commercial, industrial, retail &
4 &
Trip generation; Trip attraction; Trip distribution &
OSM \cite{geofabrik} \\
Points of interest (POI) &
Counts of POIs by class: transport, food, health, education, retail &
5 &
Trip attraction; Trip distribution &
OSM \cite{geofabrik} \\
\bottomrule
\end{tabular}
\end{table}

Population- and area-related variables describe the scale and density of residential activity. Population attributes were stratified by age and sex to capture demographic heterogeneity relevant to travel behaviour and trip generation. Employment features represent the spatial distribution of economic activity and were stratified by work type and industrial sector to reflect differences in trip attraction associated with distinct employment structures. Household characteristics were derived from Census data and stratified by household composition and car or van availability, capturing variations in mobility resources and constraints. Socioeconomic disadvantage was represented using income and employment deprivation indicators, which account for structural constraints on travel demand and accessibility. Land-use areas and POIs extracted from OSM describe functional land use and activity intensity, supporting the representation of both trip attraction and spatial interaction. The OSM keys and values used for land-use and POI extraction are documented in Supplementary Appendix~S2.

All datasets were harmonised to the 2021 LSOA geography (LSOA21). Population and household variables were already available at LSOA21. Employment data and deprivation indices originally published at the LSOA11 geography were converted using the ONS exact-fit lookup \cite{lsoalookup}. For employment data, one-to-many splits were disaggregated using population shares, and many-to-one merges were aggregated by summation to preserve total employment counts. For deprivation indices, one-to-many splits inherited the parent score, while many-to-one merges were aggregated by means. Minor boundary inconsistencies were resolved through manual spatial inspection.

Following harmonisation, all features were z-score normalised. Principal component analysis (PCA) was applied to the full feature matrix, and the top $k$ components were retained as the final feature vector for each LSOA. The resulting vectors were assigned to the nearest node in the driving network based on population-weighted LSOA centroids in the British National Grid \cite{populationweightedcentroids}, yielding the feature set 
$\mathcal{X}=\{\,x_n\in\mathbb{R}^F:n\in\mathcal{N}_{\mathrm{LSOA}}, \mathcal{N}_{\mathrm{LSOA}}\subseteq \mathcal{N}\}$,  and a node-to-LSOA mapping dictionary. Descriptive statistics and correlation analysis of the raw features are provided in the Supplementary Appendix~S3.

\subsubsection{Traffic Volumes}\label{subsubsec:gt}

Ground-truth traffic volumes were obtained from the National Highways Traffic Information System (TRIS), which provides continuous traffic monitoring on the strategic road network in England \cite{TRIS2024}. TRIS sensors record traffic flow and speed by vehicle class at 15-minute intervals.

To associate sensors with the driving network, each sensor was spatially matched to the nearest edge in the simplified UK driving network. Candidate edges were restricted to those classified as motorway, motorway link, trunk, and trunk link, which together cover the primary scope of TRIS monitoring \cite{national_highways}. For each sensor, a nearest-neighbour search was conducted among all edges intersecting a 5\,km bounding-box buffer, and the edge with the minimum perpendicular distance was selected. Sensors were retained only if the nearest distance was below 10\,m; those exceeding this threshold were excluded. The resulting sensor-to-edge mappings were visually inspected to confirm positional accuracy.

\begin{figure}[H]
    \makebox[\textwidth][l]{\hspace{-17mm}\includegraphics[width=1.2\linewidth]{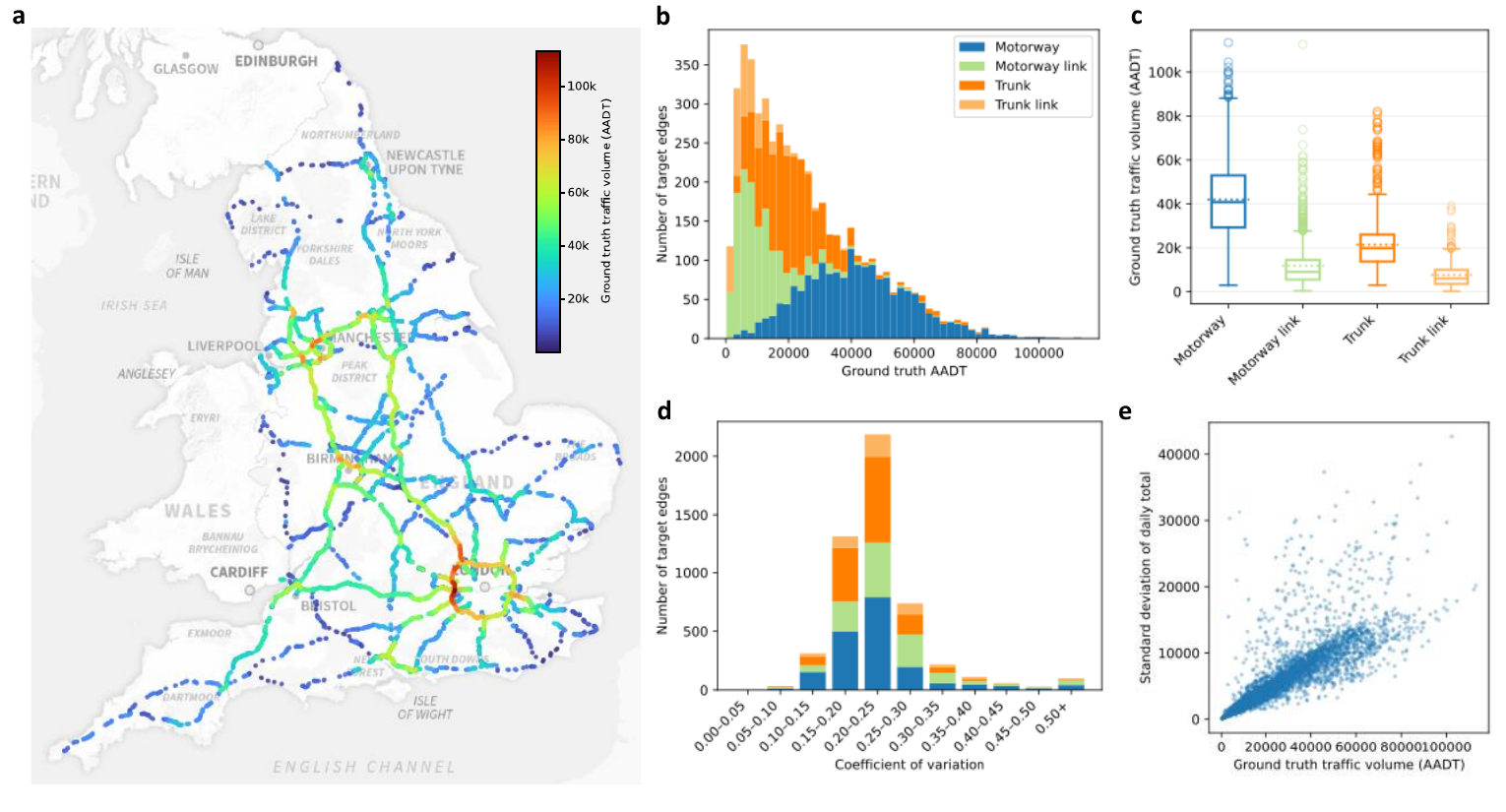}}
    \caption{\textbf{Descriptive statistics and spatial visualisation of ground-truth traffic volumes (AADT).}
\textbf{a}, Spatial distribution of AADT, with high volumes concentrated on major corridors between large cities, particularly around London.
\textbf{b},\textbf{c}, AADT distributions by highway type.
\textbf{d}, Distribution of the coefficient of variation (standard deviation divided by mean daily traffic).
\textbf{e}, Relationship between AADT and the standard deviation of daily total traffic, indicating increasing variability with higher traffic volumes.}
  \label{fig:gt_descriptive}
\end{figure}

For each retained sensor, annual average daily traffic (AADT) was computed for each year from 2017 to 2024. Sensors with fewer than 200 valid observations in a given year were excluded for that year. When multiple sensors were matched to the same edge, AADT values were aggregated using the median across sensors. For edges with valid observations in multiple years, AADT values were further aggregated using the median across years to obtain a stable long-term estimate. The number of valid years, valid days, and matched sensors per target edge are summarised in Supplementary Appendix~S4. After filtering and aggregation, a total of 5{,}088 edges were annotated with valid AADT volumes $\mathcal{Y}=\{\,y_{e^\star}\in\mathbb{R}: e^\star\in\mathcal{E}_{\mathrm{GT}}, \mathcal{E}_{\mathrm{GT}}\subseteq \mathcal{E}\}$, which served as the ground truth for model training and evaluation. Descriptive statistics and spatial visualisations of the resulting traffic volumes are shown in Figure~\ref{fig:gt_descriptive}.

\subsection{The DeepDemand Model}\label{subsec:model}

The DeepDemand model is inspired by the structure of the TDM. Its objective is to predict traffic volumes $\hat{\mathcal{Y}}$ given the driving network $\mathcal{G}$ and node-level features $\mathcal{X}$. To ensure that each prediction is informed only by information relevant to the corresponding target edge, and to allow the deep learning component to focus on feature representation and nonlinear interactions, the DeepDemand framework is organised into two sequential stages.

\subsubsection{Local OD Region Extraction and OD Pair Screening}\label{subsubsec:subgraph}

In the first stage, we identify the local spatial context that is functionally associated with each target edge $e^\star \in \mathcal{E}_{\mathrm{GT}}$. Rather than assuming that the entire national road network contributes to every prediction, we extract a local subgraph that contains only the plausible origin and destination areas whose trips are likely to traverse the target edge.

Let the target edge be $e^\star=(u,v)$, where $u$ and $v$ denote the upstream and downstream nodes, respectively. To identify potential origins, we propagate a search backward from $u$ by reversing network directionality. To identify potential destinations, we propagate a search forward from $v$ along the original network direction. However, independent forward and backward Dijkstra expansions lead to increasing overlap between the two regions as travel time grows, causing ambiguous assignment of nodes to origin or destination territories (see Figure S1.7). 

To avoid this issue, we introduce a competitive two-source Dijkstra procedure, summarised in Algorithm~\ref{alg:twosource}. The origin-side and destination-side searches are initiated simultaneously and executed within a shared priority queue ordered by travel time. When a node is reached, it is permanently assigned to the side that arrives first and excluded from further expansion by the opposing search. This competitive mechanism ensures that the resulting origin and destination regions remain mutually exclusive and form a clear partition of the local network.

\begin{figure}[htbp]
    \makebox[\textwidth][l]{\hspace{-10mm}\includegraphics[width=1.1\linewidth]{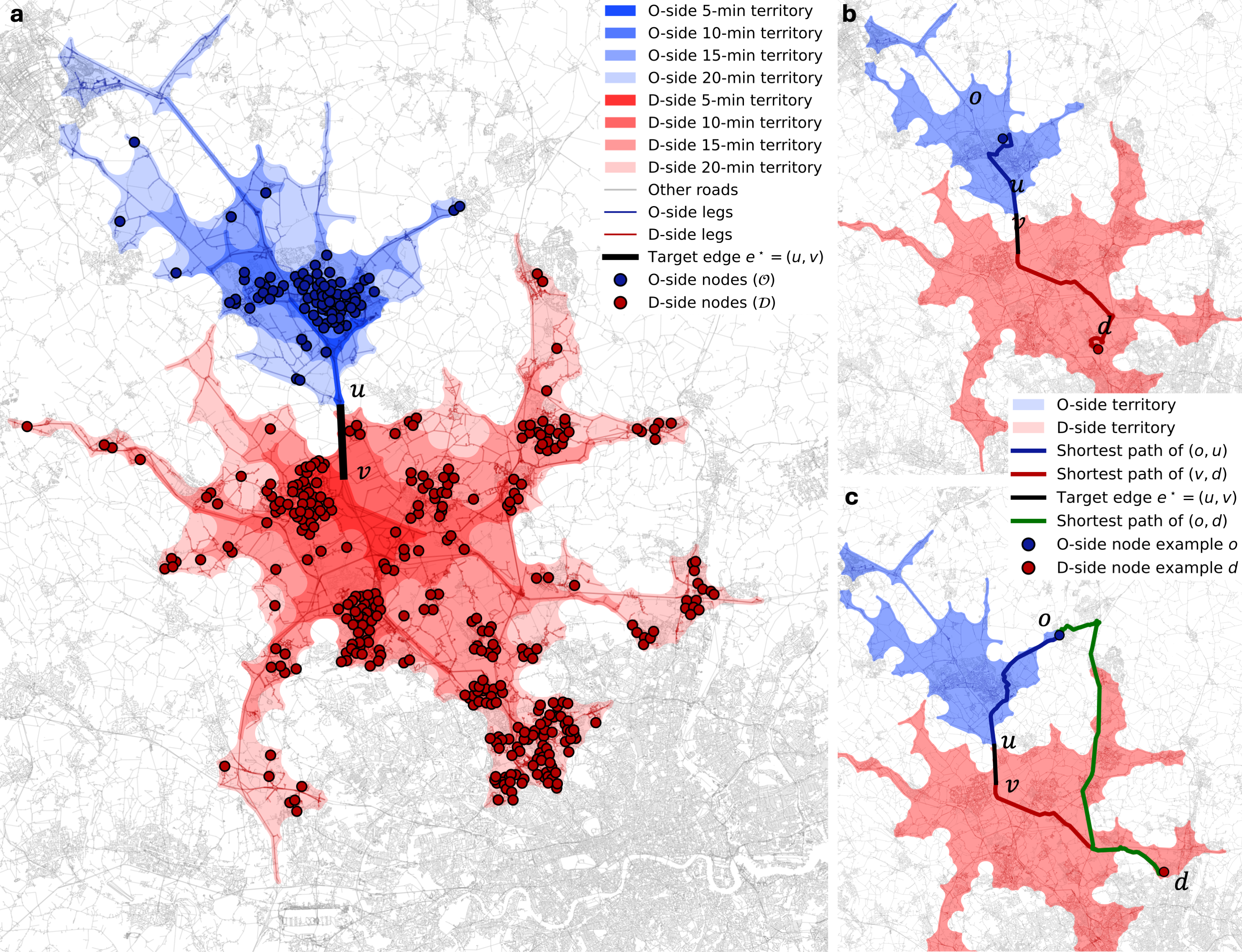}}
\caption{\textbf{Local OD region extraction and OD pair screening.} \textbf{a}, Illustration of the competitive expansion process from the target edge, shown up to a 20-minute travel-time threshold for simplicity; points indicate potential origin and destination locations (LSOA centroids) identified during the search. \textbf{b}, Example of a valid OD pair whose shortest route necessarily traverses the target edge. \textbf{c}, Example of an invalid OD pair for which a faster bypass route exists, leading to exclusion.}
  \label{fig:isochrone}
\end{figure}

During expansion, all visited nodes carrying LSOA features are recorded, together with their shortest travel times from the target edge. This process yields two node sets for each target edge, $\mathcal{O} \subseteq \mathcal{N}$ and $\mathcal{D} \subseteq \mathcal{N}$, representing feasible upstream origins and downstream destinations of trips that pass through $e^\star$. A travel-time cutoff of one hour is applied in both directions. The extraction process is illustrated in Figure~\ref{fig:isochrone}a.

\begin{algorithm}[htbp]
\caption{\textbf{Competitive two-source Dijkstra procedure}}
\label{alg:twosource}
\small
\begin{algorithmic}[1]
\Require MultiDiGraph $G=(\mathcal{N},\mathcal{E})$, target edge $e^\star=(u,v)$, travel-time cutoff $\tau$
\State Initialise priority queue 
       $Q \gets \{(0,\mathrm{O},u),(0,\mathrm{D},v)\}$ \Comment{$\mathrm{O}$: origin-side, $\mathrm{D}$: destination-side}
\State Initialise travel-time maps $t_{\mathrm{O}}(i)\gets\infty$, $t_{\mathrm{D}}(i)\gets\infty$ for all $i\in\mathcal{N}$
\State Set $t_{\mathrm{O}}(u)\gets 0$, $t_{\mathrm{D}}(v)\gets 0$
\State Initialise predecessor maps $p_{\mathrm{O}}(i)\gets\bot$, $p_{\mathrm{D}}(i)\gets\bot$ for all $i\in\mathcal{N}$
\State Initialise winner map $w(i)\gets\bot$ for all $i\in\mathcal{N}$
\While{$Q$ not empty}
    \State $(t,s,i) \gets$ pop minimum from $Q$ \Comment{$s \in \{\mathrm{O},\mathrm{D}\}$, $i$ current node}
    \If{$t > \tau$ \textbf{or} $w(i) \neq \bot$} \State \textbf{continue} \EndIf
    \State $w(i) \gets s$ \Comment{first-arrival claims node $i$ for side $s$}
    \For{each neighbour $j$ of $i$ allowed for side $s$}
        \If{$w(j) \neq \bot$}
            \State \textbf{continue} \Comment{node $j$ already claimed and final}
        \EndIf
        \State $t_{ij} \gets \mathrm{travel\_time}(i,j)$
        \State $t' \gets t + t_{ij}$
        \If{$t' \le \tau$}
            \If{$s = \mathrm{O}$ \textbf{and} $t' < t_{\mathrm{O}}(j)$}
                \State $t_{\mathrm{O}}(j) \gets t'$
                \State $p_{\mathrm{O}}(j) \gets i$
                \State push $(t', \mathrm{O}, j)$ into $Q$
            \ElsIf{$s = \mathrm{D}$ \textbf{and} $t' < t_{\mathrm{D}}(j)$}
                \State $t_{\mathrm{D}}(j) \gets t'$
                \State $p_{\mathrm{D}}(j) \gets i$
                \State push $(t', \mathrm{D}, j)$ into $Q$
            \EndIf
        \EndIf
    \EndFor
\EndWhile
\State $\mathcal{O} \gets \{\, i\in\mathcal{N} : w(i)=\mathrm{O}\,\}$
\State $\mathcal{D} \gets \{\, i\in\mathcal{N} : w(i)=\mathrm{D}\,\}$
\State \Return $\mathcal{O},\mathcal{D},\, t_{\mathrm{O}},t_{\mathrm{D}}$ \textbf{(optionally also $p_{\mathrm{O}},p_{\mathrm{D}}$)}\end{algorithmic}
\end{algorithm}

After constructing $\mathcal{O}$ and $\mathcal{D}$, we further screen all candidate OD pairs $(o,d) \in \mathcal{O} \times \mathcal{D}$ to retain only those that necessarily traverse the target edge. Let $t_{\mathrm{O}}(o)$ denote the shortest travel time from $o \in \mathcal{O}$ to $u$, $t_{\mathrm{D}}(d)$ the shortest travel time from $v$ to $d \in \mathcal{D}$, and $t_{e^\star}$ the travel time along the target edge $(u,v)$. For each candidate pair, we compute the shortest travel time $t_{\mathrm{OD}}(o,d)$ on the full network without constraining the route. If $t_{\mathrm{O}}(o) + t_{e^\star} + t_{\mathrm{D}}(d) = t_{\mathrm{OD}}(o,d)$, then the shortest-time route from $o$ to $d$ passes through $e^\star$, and the OD pair is retained (Figure~\ref{fig:isochrone}b). Otherwise, a faster bypass exists, and the pair is discarded (Figure~\ref{fig:isochrone}c). This deterministic criterion ensures that only OD pairs whose fastest routes necessarily use the target edge are included. 

The combined extraction and screening procedure yields a local set of valid OD pairs for each target edge,
\begin{equation}
\mathcal{P}^\star = \bigl\{\, (o,d) \in \mathcal{O} \times \mathcal{D} \;\big|\; t_{\mathrm{O}}(o) + t_{e^\star} + t_{\mathrm{D}}(d) = t_{\mathrm{OD}}(o,d) \,\bigr\},
\end{equation}
together with their associated travel times. These outputs form the inputs to the subsequent deep learning stage.

\subsubsection{Deep Learning Component}\label{subsubsec:dl}

\begin{figure}[htbp]
    \makebox[\textwidth][l]{\hspace{+30mm}\includegraphics[width=0.7\linewidth]{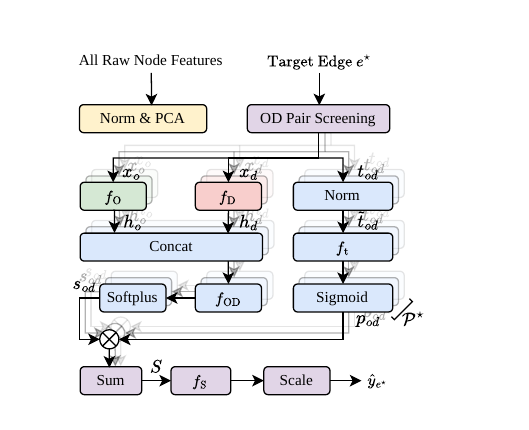}}
  \caption{\textbf{The architecture of the deep learning model.}}
  \label{fig:DL}
\end{figure}

The second stage of the DeepDemand model estimates the traffic volume on each target edge $e^\star$ by aggregating contributions from all valid OD pairs. The architecture of the deep learning component is shown in Figure~\ref{fig:DL}. Each OD pair $(o,d)$ contributes to the predicted volume through three elements: the trip-generation strength at the origin, the trip-attraction strength at the destination, and a travel-cost deterrence term.

For each valid OD pair, the corresponding LSOA feature vectors $x_o$ and $x_d$ are retrieved from the feature bank $\mathcal{X}$. Two independent learnable functions encode these vectors into origin and destination embeddings,
\begin{equation}
h_o = f_{\mathrm{O}}(x_o),\qquad h_d = f_{\mathrm{D}}(x_d),
\end{equation}
which represent latent propensities for trip generation and attraction at $o$ and $d$, respectively. The encoders do not share parameters, allowing the model to capture directional asymmetries between origins and destinations.

The origin and destination embeddings are then concatenated and passed through a learnable interaction function to produce a non-negative OD pair score,
\begin{equation}
s_{od} = \phi\!\left(f_{\mathrm{OD}}([h_o\,\|\,h_d])\right), \qquad \phi = \text{Softplus},
\end{equation}
which represents the total flow potential between $o$ and $d$ in the absence of travel cost. The Softplus activation enforces non-negativity, analogous to a trip-distribution weight.

Travel cost is incorporated through a learnable deterrence function applied to the OD travel time,
\begin{equation}
p_{od} = \sigma\left( f_{\mathrm{t}} (\tilde{t}_{od})\right),
\quad
\tilde{t}_{od} = \frac{t_{od} - \mu}{s},
\end{equation}
where $\mu$ and $s$ are global normalisation constants estimated across all valid OD pairs, and $\sigma$ denotes the sigmoid function. The resulting value $p_{od}\in(0,1)$ can be interpreted as the fraction of the total flow potential $s_{od}$ that remains after accounting for travel impedance.

The predicted AADT for the target edge is obtained by aggregating contributions across all valid OD pairs,
\begin{equation}
S = \sum_{(o,d)\in \mathcal{P}^\star} s_{od}\, p_{od},
\qquad
\hat{y}_{e^\star} = \gamma f_{\mathrm{S}}(S),
\end{equation}
where $\gamma$ is a fixed scaling constant and $f_{\mathrm{S}}$ is a learnable monotonic transformation that maps the model-implied free-flow volume $S$ to the observed equilibrium traffic volume.

Under this formulation, the node encoders correspond to trip-generation and trip-attraction components, the OD interaction module performs trip distribution, and the deterrence function learns a data-driven impedance relationship. The final aggregation implicitly performs traffic assignment by summing only over OD pairs whose shortest paths traverse the target edge, while absorbing congestion, capacity constraints, and other equilibrium effects into the learned transformation. All components are differentiable and are trained jointly using observed traffic volumes.

\subsubsection{Training Setup}\label{subsec:train}

Model performance is assessed using both random five-fold cross-validation and spatial cross-validation by region. Random cross-validation evaluates overall predictive accuracy, while spatial cross-validation explicitly tests the model’s ability to transfer to regions that are not observed during training and mitigates potential optimism arising from spatial autocorrelation between neighbouring segments. The baseline models comprise ordinary least squares (OLS) linear regression, ridge regression with L2 regularisation, a Random Forest regressor, and a classical gravity-style interaction model. Full methodological details are provided in Supplementary Appendix S5.

Training is conducted in a stochastic, edge-centric manner. At each iteration, a single target edge is sampled from the training set, and all inputs associated with this edge are assembled, including its valid OD pairs, corresponding travel times, and LSOA features. A forward pass produces a traffic volume prediction for the target edge, after which the prediction error is computed and back-propagated to update model parameters. Parameters are updated after each target-edge sample, corresponding to stochastic gradient descent on the expected prediction error.

Let $\mathcal{E}_{\text{set}}$ denote the set of target edges in the current dataset (training or test), and let $e^{\star} \in \mathcal{E}_{\text{set}}$ denote a specific target edge. The observed and predicted traffic volumes on $e^{\star}$ are denoted by $y_{e^{\star}}$ and $\hat{y}_{e^{\star}}$, respectively.

The training objective minimises the squared error for a single target edge,
\begin{equation}
\mathcal{L}_{\text{train}} =
\left( y_{e^{\star}} - \hat{y}_{e^{\star}} \right)^2 ,
\end{equation}
which provides an unbiased stochastic estimate of the mean squared error over the training set.

Model performance is evaluated using multiple complementary metrics. The GEH statistic for a target edge is defined as
\begin{equation}
\text{GEH}_{e^{\star}} =
\sqrt{2 \,\frac{\left( y_{e^{\star}} - \hat{y}_{e^{\star}} \right)^2}{y_{e^{\star}} + \hat{y}_{e^{\star}}}} ,
\end{equation}
and the mean GEH across a dataset is
\begin{equation}
\text{MGEH} =
\frac{1}{\lvert \mathcal{E}_{\text{set}} \rvert}
\sum_{e^{\star} \in \mathcal{E}_{\text{set}}}
\text{GEH}_{e^{\star}} .
\end{equation}

The mean absolute error (MAE) is computed as
\begin{equation}
\text{MAE} =
\frac{1}{\lvert \mathcal{E}_{\text{set}} \rvert}
\sum_{e^{\star} \in \mathcal{E}_{\text{set}}}
\left| y_{e^{\star}} - \hat{y}_{e^{\star}} \right| ,
\end{equation}
and the coefficient of determination is given by
\begin{equation}
R^2 =
1 -
\frac{
\sum_{e^{\star} \in \mathcal{E}_{\text{set}}}
\left( y_{e^{\star}} - \hat{y}_{e^{\star}} \right)^2
}{
\sum_{e^{\star} \in \mathcal{E}_{\text{set}}}
\left( y_{e^{\star}} - \bar{y} \right)^2
},
\end{equation}
where $\bar{y}$ denotes the mean observed traffic volume over $\mathcal{E}_{\text{set}}$.

Unless otherwise stated, the default configuration uses multilayer perceptrons (MLPs) for $f_{\mathrm{O}}$, $f_{\mathrm{D}}$, $f_{\mathrm{OD}}$, and $f_{\mathrm{t}}$. The node encoders $f_{\mathrm{O}}$ and $f_{\mathrm{D}}$ use hidden and output dimensions of $[16,16]$, the OD pair scorer $f_{\mathrm{OD}}$ uses dimensions $[16,8]$, and the travel-time deterrence network $f_{\mathrm{t}}$ uses dimensions $[16,16]$. Travel times are standardised using global normalisation parameters $\mu = 3600\,\text{s}$ and $\sigma = 1000\,\text{s}$. LSOA feature vectors are reduced to $k=64$ dimensions using PCA. In the final aggregation step, the scaling constant is set to $\gamma=100$, and a square-root function is used for $f_{\mathrm{S}}$. Early stopping is applied with a patience of 20 evaluation steps and a minimum improvement threshold of 0.1 in MGEH, with model evaluation performed every 1,000 training iterations.

Training uses the AdamW optimiser with a learning rate of 0.001, weight decay of $10^{-4}$, and gradient norm clipping at 5 to ensure numerical stability. All experiments are conducted on a local system (Windows~11) equipped with 16\,GB RAM and a single GPU (16\,GB VRAM), using Python~3.10 and PyTorch~2.5.1 with CUDA~12.1. The OD pair screening procedure is computationally intensive and is therefore executed once in parallel across target edges on high-performance computing clusters.

\section{Results}\label{subsec:results}
\subsection{Model Evaluation}

The computational requirements of DeepDemand are moderate and compatible with routine large-scale deployment. Training a single model for one random cross-validation fold requires approximately 14 hours under our computational setting, while inference for a single target edge takes about 0.25 seconds. The most computationally intensive component is the local OD region extraction and OD pair screening procedure, which requires approximately 19~GB of memory and around 2 hours per target edge. Importantly, this step is performed only once, can be fully parallelised across high-performance computing resources, and is reused across all subsequent training and inference stages. As a result, it does not introduce additional computational overhead during model fitting or prediction.

Observed AADT values across all target edges exhibit substantial heterogeneity, with a maximum of 113,436 vehicles, a mean of 25,243, and a standard deviation of 18,893. Daily traffic volumes also show considerable variability, with a mean coefficient of variation of 0.24 (Figure~\ref{fig:gt_descriptive}d). Despite this wide dynamic range, DeepDemand achieves strong and stable predictive performance, as shown in Table~\ref{tab:eval}. Across random five-fold cross-validation, the model attains a mean MGEH of 54.31, an MAE of 7,406 vehicles, and an $R^2$ of 0.718. These results indicate that the model accurately captures absolute traffic magnitudes and their relative variation across the national network, and they consistently outperform all baseline benchmark models under the same evaluation protocol.

\begin{table}[ht]
\centering
\small
\caption{\textbf{Baseline comparison under random five-fold and spatial cross-validation.} Values are reported as mean (standard deviation).}
\label{tab:eval}
\begin{tabular}{lcccccc}
\toprule
& \multicolumn{3}{c}{Train} & \multicolumn{3}{c}{Test} \\
\cmidrule(lr){2-4} \cmidrule(lr){5-7}
Model & MGEH & MAE & $R^2$ & MGEH & MAE & $R^2$ \\
\midrule
\multicolumn{7}{l}{\textbf{Random five-fold CV}} \\
Linear regression      & 85.09 (0.36) & 13326 (67) & 0.204 (0.006) & 89.42 (0.74) & 14255 (126) & -0.001 (0.124) \\
Ridge regression       & 85.11 (0.39) & 13332 (69) & 0.203 (0.006) & 89.21 (0.62) & 14208 (117) & 0.012 (0.119) \\
Random forest          & 26.01 (0.17) & 3441 (16)  & 0.934 (0.001) & 56.03 (0.74) & 8389 (62)   & 0.641 (0.014) \\
Gravity (log-linear)   & 65.45 (0.22) & 9374 (51)  & 0.549 (0.005) & 65.60 (0.31) & 9399 (99)   & 0.547 (0.016) \\
DeepDemand             & 50.26 (0.67) & 6838 (70)  & 0.746 (0.009) & 54.31 (1.13) & 7406 (166)  & 0.718 (0.019) \\
\midrule
\multicolumn{7}{l}{\textbf{Spatial CV}} \\
Linear regression      & 85.14 (0.90) & 13325 (223) & 0.199 (0.005) & 92.49 (11.10) & 14813 (3186) & -0.096 (0.190) \\
Ridge regression       & 85.15 (0.91) & 13328 (224) & 0.198 (0.005) & 92.34 (11.25) & 14780 (3215) & -0.084 (0.182) \\
Random forest          & 25.38 (0.26) & 3340 (47)   & 0.936 (0.003) & 65.70 (8.57)  & 10211 (2021) & 0.452 (0.077) \\
Gravity (log-linear)   & 65.41 (0.65) & 9363 (139)  & 0.545 (0.012) & 66.81 (6.52)  & 9565 (1636)  & 0.478 (0.188) \\
DeepDemand             & 51.09 (0.85) & 7035 (185)  & 0.736 (0.016) & 56.43 (4.36)  & 7669 (1171)  & 0.665 (0.092) \\
\bottomrule
\end{tabular}
\end{table}

\begin{figure}[htbp]
    \makebox[\textwidth][l]{\hspace{-5mm}\includegraphics[width=1.\linewidth]{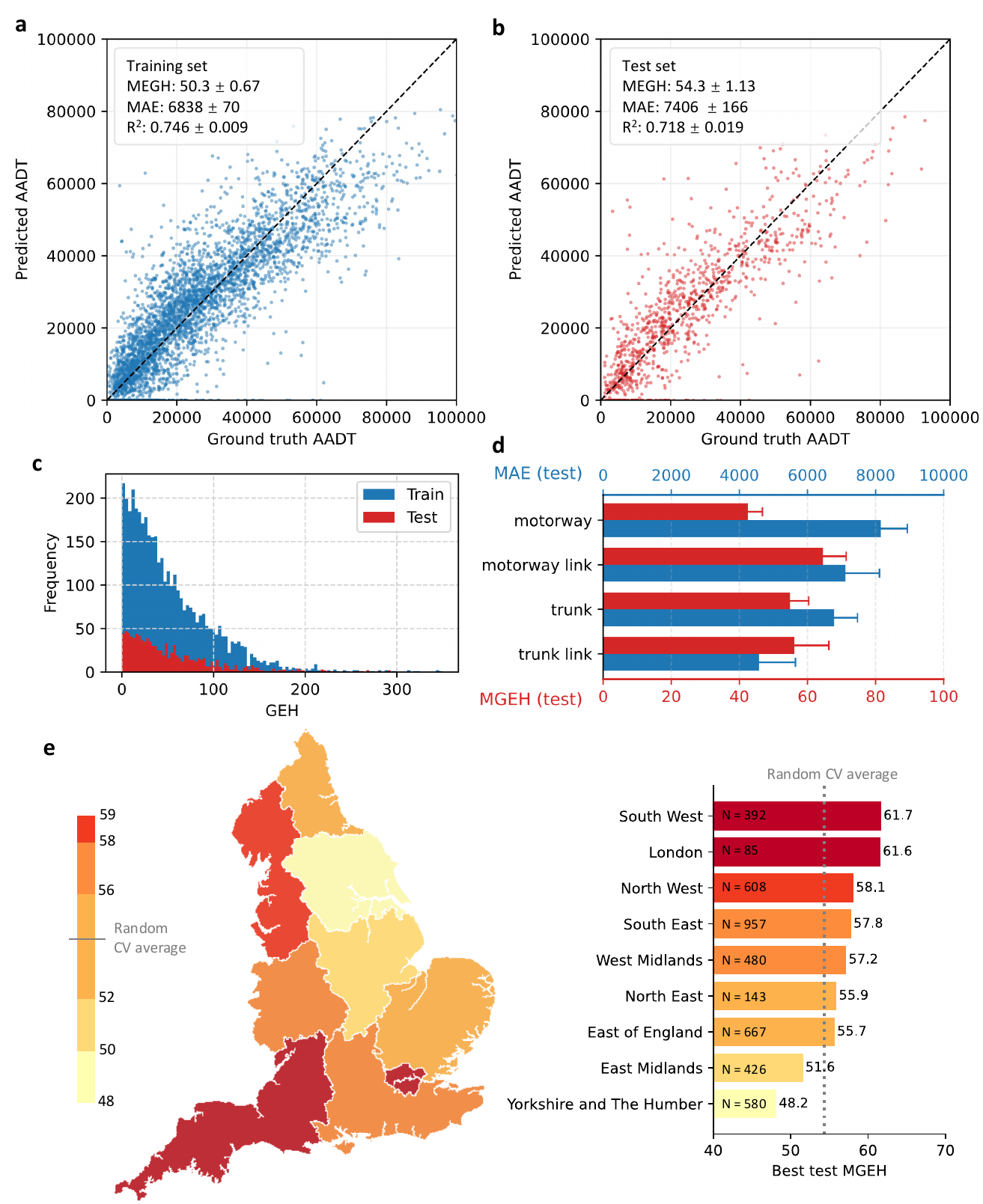}}
    \caption{\textbf{Model evaluation under random five-fold cross-validation.} Reported metrics show the mean and standard deviation across folds. All panels visualise results from the first fold. \textbf{a}, Predicted versus observed AADT for the training set; \textbf{b}, predicted versus observed AADT for the test set; \textbf{c}, distribution of the GEH metric across all target edges; \textbf{d}, MGEH and MAE stratified by road type of the target edge; \textbf{e}, Result of spatial cross-validation.}
  \label{fig:eval}
\end{figure}

Scatter plots of predicted versus observed AADT for both training and test sets (Figure~\ref{fig:eval}a,b) show a tight alignment along the identity line across the full range of traffic volumes, including both low-volume and high-volume roads. The high similarity between training and test performance indicates limited overfitting and strong generalisation capacity. The distribution of GEH values (Figure~\ref{fig:eval}c) further confirms that the majority of target edges fall within acceptable error ranges, with no evidence of systematic bias. Prediction accuracy is consistent across different functional road classes, including both main roads and link roads (Figure~\ref{fig:eval}d). This demonstrates the model’s ability to distinguish geographically proximate yet semantically distinct road segments, such as slip roads and adjacent arterial links. This capability is critical for network-wide transferability. 

Spatial cross-validation further demonstrates the robustness of the model under strict geographic transfer conditions. Under spatial CV, the model attains a mean MGEH of 56.43, an MAE of 7,669 vehicles, and an $R^2$ of 0.665, closely matching the performance observed under random cross-validation. Prediction performance varies only modestly across English regions and consistently fluctuates around the random cross-validation average (Figure~\ref{fig:eval}e). No region exhibits extreme degradation in performance, indicating that the model generalises well to unseen spatial contexts. Slightly higher errors are observed in London and the South West, likely reflecting the distinct traffic regimes at the two extremes of the volume distribution: very high flows in metropolitan London and generally low flows in the South West. Importantly, these deviations remain limited and do not compromise overall model stability.

Further analyses indicate that prediction accuracy is slightly higher for target edges with mid-to-high traffic volume deciles. However, model performance is largely insensitive to data availability characteristics, including the number of valid observation years and the number of sensors associated with each target edge (Supplementary Appendix~S6.1). Spatial visualisation of residuals reveals no systematic spatial clustering or directional bias (Supplementary Appendix~S6.2), suggesting that remaining errors are predominantly stochastic rather than structural.

\subsection{Model Explainability}

\subsubsection{Travel-time Deterrence Function}\label{subsubsec:deterrence}

We compare several candidate specifications for the travel-time deterrence function $f_{\mathrm{t}}$, including a linear logit (sigmoid) form with learnable parameters, an exponential decay function with a learnable rate, and an MLP. The MLP consistently outperforms the parametric alternatives in predictive accuracy. By avoiding a predefined functional shape, it enables the deterrence relationship to be inferred flexibly from the data.

To interpret the learned deterrence behaviour, we visualise the MLP-based $f_{\mathrm{t}}$ for each cross-validation fold. Specifically, the trained MLP parameters are extracted and a forward pass is performed over a dense grid of OD travel times $t_{od} \in [0, 120]$ minutes. The resulting curve represents the relationship between travel time and $p_{od}$, defined as the ratio of actual OD flow to total flow potential.

\begin{figure}[htbp]
    \makebox[\textwidth][l]{\hspace{+5mm}\includegraphics[width=0.9\linewidth]{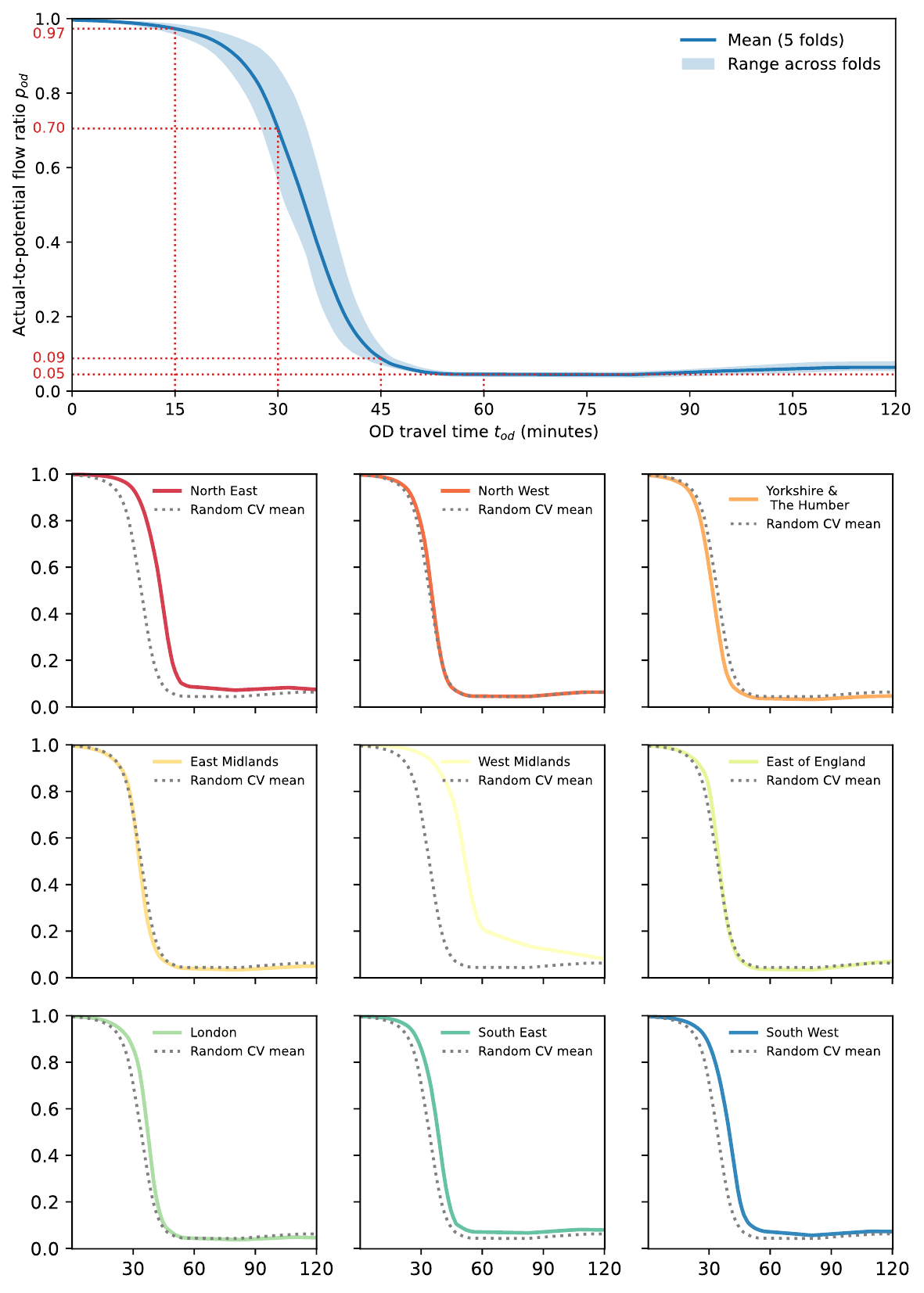}}
\caption{\textbf{Learned travel-time deterrence function.} The upper panel shows the mean curve from random five-fold cross-validation with pointwise variability across folds. The lower panels show fold-specific curves from spatial cross-validation, each corresponding to a different held-out region.}  \label{fig:time}
\end{figure}

Figure~\ref{fig:time} shows the mean deterrence curve obtained by averaging the fold-specific predictions at each value of $t_{od}$, with the shaded band indicating the pointwise range across folds. The narrow uncertainty band demonstrates that the learned deterrence function is highly stable under cross-validation. For travel times below 15 minutes, $p_{od}$ declines only marginally. Between 15 and 45 minutes, $p_{od}$ decreases sharply, dropping from 0.97 to 0.09, indicating a rapid decay in car-use likelihood with increasing travel time. Beyond 45 minutes, $p_{od}$ stabilises at a low level, suggesting that car trips of this duration or longer are relatively rare. Notably, $p_{od}$ does not converge to zero. Instead, it plateaus at around 0.05 after 50 minutes and even exhibits a slight increase to approximately 0.07 as travel time extends from 90 to 120 minutes. This tail behaviour indicates the persistent presence of long-duration car journeys, albeit at low probability. Across the nine spatial cross-validation folds, the learned deterrence functions exhibit a consistent overall shape but with small shifts, suggesting modest regional variation in travel-time sensitivity and underlying travel behaviour.

\subsubsection{Explaining OD pair scores using LSOA features} \label{subsubsec:shap}

Given the explicitly factorised and interpretable structure of the proposed model, it is informative to examine which origin–destination (OD) characteristics are associated with higher OD pair scores ($s_{od}$). We first inspect the learned LSOA embeddings using UMAP, which reveal coherent large-scale structure and smooth spatial transitions (Supplementary Appendix~S7). We then perform a post-hoc feature attribution analysis using SHAP values to quantify how individual LSOA features influence OD pair scores.

We construct the full universe of unique OD pairs across all target edges, resulting in 107,317,306 distinct OD pairs. From this pool, we randomly sample 50,000 OD pairs and recompute their OD scores using the trained model and the same PCA transformations applied during training. The logarithm of the OD pair score is then used as the dependent variable, while raw LSOA-level features at the origin and destination are used as explanatory variables. To obtain robust and stable attributions, we fit random forest regressors and compute SHAP values as a standard post-hoc interpretability pipeline. Importantly, we train separate random forest models for each feature group (e.g., population age, employment sector, land use, POIs), rather than pooling all features into a single model. This design avoids confounding effects from nested or highly correlated variables and ensures that SHAP attributions remain meaningful and comparable across feature strata.

\begin{figure}[htbp]
    \makebox[\textwidth][l]{\hspace{-20mm}\includegraphics[width=1.2\linewidth]{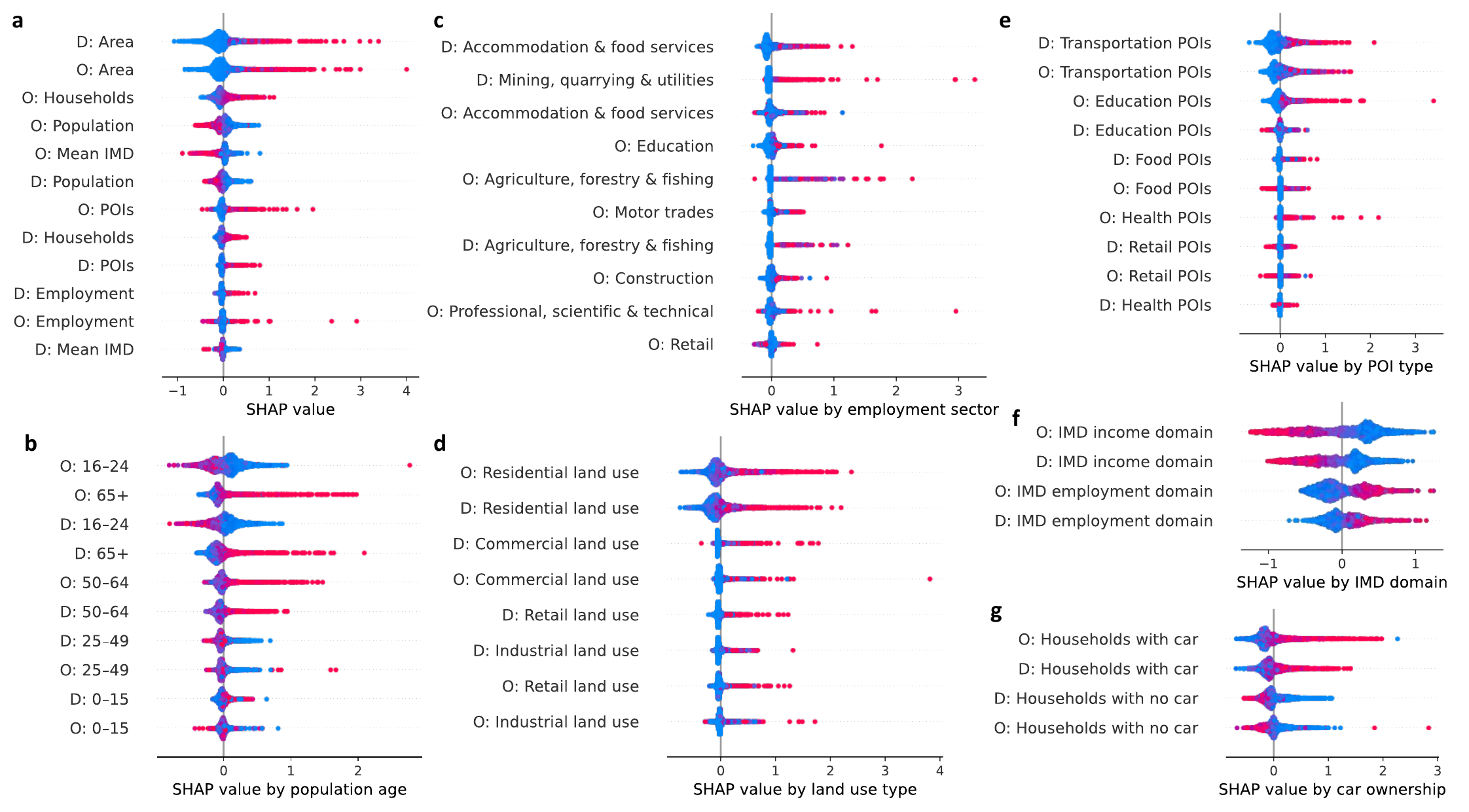}}
\caption{\textbf{SHAP value distributions illustrating the influence of origin and destination LSOA characteristics on OD pair scores.} SHAP values are computed from random forest models fitted using OD pair scores as the response variable and grouped LSOA-level features as predictors. Panels correspond to different feature groups: \textbf{a}, aggregated first-level feature domains, \textbf{b}, population age structure, \textbf{c}, employment sector composition, \textbf{d}, land-use composition by area, \textbf{e}, POI counts by type, \textbf{f}, deprivation domains, and \textbf{g}, household car ownership. Colour indicates feature magnitude, with red representing higher values and blue lower values.}
  \label{fig:shap}
\end{figure}

The results are summarised in Figure~\ref{fig:shap}. At the coarsest level (Figure~\ref{fig:shap}a), the geographical area of both origin and destination LSOAs emerges as the most influential determinant of OD pair scores. Larger LSOAs are consistently associated with higher scores. This reflects a structural effect: in geographically large LSOAs, the number of possible OD alternatives is smaller, requiring individual OD pairs to carry greater weight to aggregate to the observed traffic volume on the target edge. Following area, the number of households at the origin is a strong positive contributor, while total population shows an inverse relationship with OD scores. 

Disaggregating these high-level effects reveals further behavioural insights. For population age structure (Figure~\ref{fig:shap}b), a higher number of residents aged 16–24 at either the origin or destination strongly suppresses OD scores, whereas larger populations aged 65+ and 50–64 are associated with substantially higher scores. Employment composition also plays an important role (Figure~\ref{fig:shap}c): OD pairs involving LSOAs with a high concentration of accommodation and food services employment exhibit markedly higher scores at both the origin and destination. Land-use composition shows a similarly interpretable structure (Figure~\ref{fig:shap}d), with residential land use at both the origin and destination emerging as the strongest positive contributor, followed by commercial land use. This indicates that OD pairs linking residential and mixed-use areas dominate highway demand. POI features further reinforce these findings (Figure~\ref{fig:shap}e). Transportation-related POIs at either end of an OD pair are the strongest POI-level predictors of high scores, followed by education- and food-related POIs, highlighting the role of multimodal hubs and trip-attracting destinations in shaping highway usage. Socioeconomic indicators also exhibit systematic effects: lower income deprivation and higher employment deprivation are associated with increased OD scores (Figure~\ref{fig:shap}f), while higher rates of household car ownership strongly increase OD scores and households without cars strongly suppress them (Figure~\ref{fig:shap}g). Additional SHAP analyses resolving finer-grained feature strata are provided in Supplementary Appendix~S8. These include employment stratified by full-time and part-time status, employment stratified by both sector and by contract type, population counts stratified by sex and age group, and households by household composition categories. 

\subsubsection{Identifying high-potential OD areas}
\label{subsubsec:od_potential}

Beyond explaining variation in OD pair scores through raw LSOA features, the explicit structure of the DeepDemand model also enables direct visualisation of spatial patterns in trip generation and attraction potential. This is achieved by aggregating the learned OD pair scores and attributing them back to their corresponding origin and destination areas.

Specifically, for each LSOA, we define its trip generation potential (O potential) as the average OD pair score between that LSOA acting as the origin and all other LSOAs acting as destinations. Conversely, the trip attraction potential (D potential) of an LSOA is defined as the average OD pair score between all other LSOAs acting as origins and that LSOA acting as the destination. By normalising each potential by the total number of possible OD counterparts, these definitions ensure that origin and destination potentials are directly comparable across space and are not biased by differences in the number of OD pair combinations. In addition, both potentials are further standardised by dividing by LSOA land area, yielding spatially meaningful density-based measures.

\begin{figure}[htbp]
    \makebox[\textwidth][l]{\hspace{-0mm}\includegraphics[width=1\linewidth]{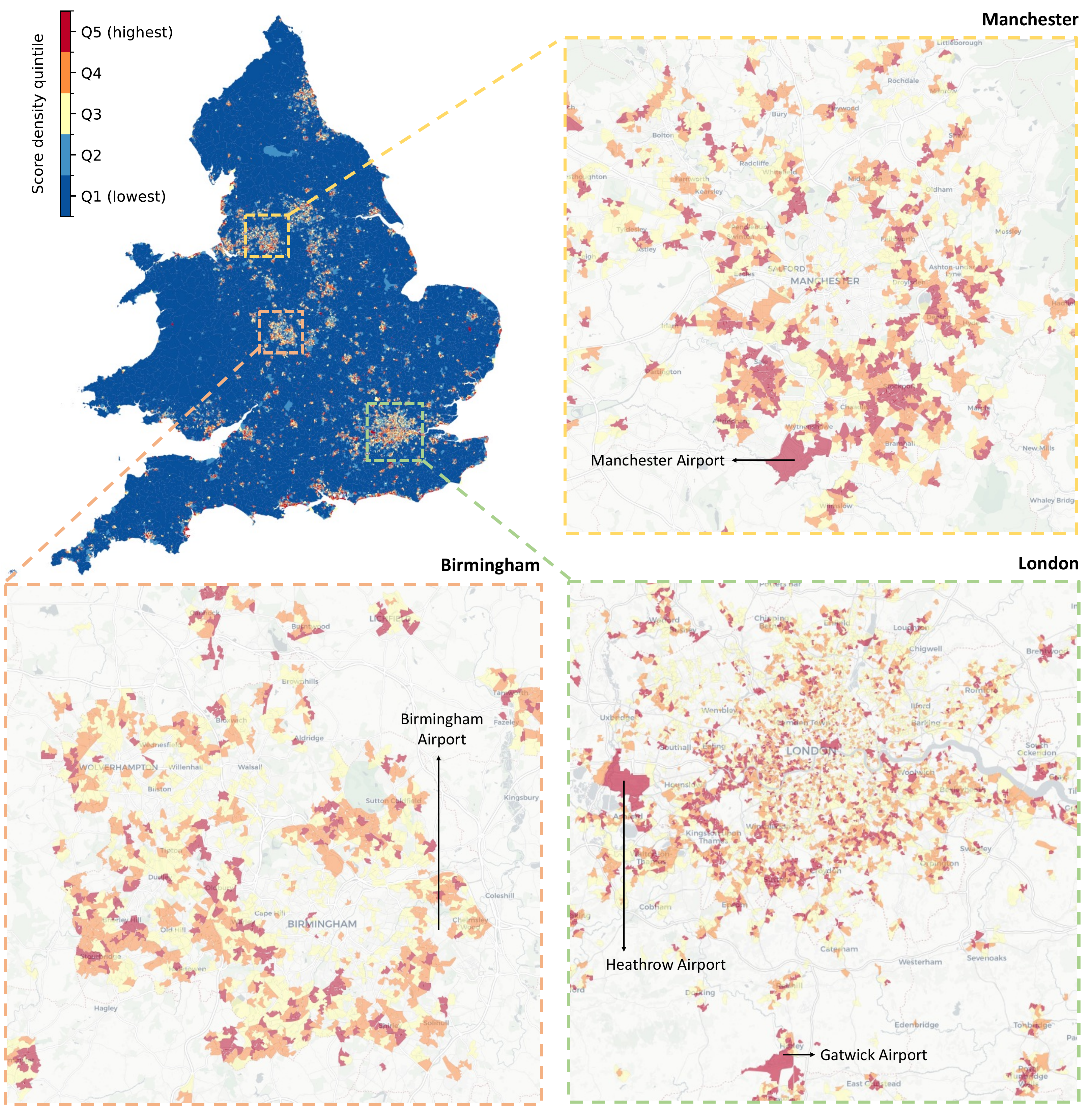}}
\caption{\textbf{Spatial distribution of standardised destination potential across LSOAs, shown as quintiles of D-potential density.} The national map highlights the overall spatial structure of trip attraction, while zoomed-in panels illustrate detailed patterns for Greater Manchester, Birmingham, and Greater London. High D-potential clusters align with major employment centres, transport hubs, and activity corridors, including key airport locations, illustrating the polycentric organisation of travel attraction recovered by the model.}
\label{fig:d_score}
\end{figure}

The spatial distribution of standardised D potential is shown in Figure~\ref{fig:d_score}. At the national scale, clear and interpretable patterns emerge. High D-potential areas align closely with major employment centres, transport hubs, and activity corridors, forming a distinctly polycentric structure rather than a single dominant core. Across England and the South East, elevated D potential is concentrated in economically active regions and along major transport axes, while predominantly residential and low-activity areas consistently exhibit low values. This spatial structure indicates that the learned destination potential captures functional urban and regional organisation, rather than merely reflecting population density.

The city-level zoom-ins further demonstrate the interpretability of the learned destination (D) potential and reveal meaningful contrasts across metropolitan regions. In Greater London, the highest D-potential clusters are concentrated in the City of London and Canary Wharf, with additional elevated areas distributed along the Thames corridor and around major transport nodes. Heathrow and Gatwick airports both emerge as prominent high-potential destinations. In Greater Manchester, Manchester Airport similarly appears as a pronounced D-potential hotspot. These patterns are consistent with the role of these airports as major employment centres and international gateways serving large passenger volumes, with access that remains strongly reliant on highway-based travel despite the availability of public transport alternatives. In contrast, Birmingham Airport does not exhibit a comparable concentration of D potential. This difference is consistent with both its substantially lower passenger throughput relative to Heathrow, Gatwick, and Manchester, and its comparatively higher share of public transport access~\cite{CAA_PassengerSurvey, DJS_AirportAccess}. 

The spatial distribution of O potential is provided in the Supplementary Appendix S9. Compared with D potential, O potential is less spatially concentrated and extends further into suburban and peri-urban areas. This pattern is intuitively consistent with the spatial structure of trip generation, which is more widely distributed across residential zones, whereas trip attraction is concentrated in a smaller number of functional centres.

\section{Discussion}\label{sec:discussion}
This study proposes DeepDemand as a data-driven alternative to traditional four-step TDMs for highway traffic volume prediction. Using only external socioeconomic and spatial context variables, the model achieves high predictive accuracy at national scale while remaining interpretable and transferable. Unlike conventional TDM pipelines, DeepDemand does not require explicit trip tables, calibrated behavioural parameters, or mode-specific assumptions, yet it recovers coherent and behaviourally meaningful traffic patterns directly from observed data. This positions DeepDemand as a complementary modelling framework that bridges classical transport theory and modern data-driven inference.

DeepDemand demonstrates strong and stable predictive performance across England under both random and spatial cross-validation. Random cross-validation reached a mean $R^2$ of 0.718 and a mean MAE of 7406, with spatial cross-validation shows comparable stability and limited regional degradation, indicating robust generalisation across geographic contexts. Compared with existing studies targeting similar outcomes, DeepDemand represents a substantial advance in both scale and scope. For example, Das and Tsapakis~\cite{das2020} reported a mean $R^2$ of 0.36 when predicting AADT on low-volume roads using census and survey data, while Ganji et al.~\cite{ganji2022} achieved $R^2 = 0.58$ using aerial imagery for urban roads with substantially higher error. Studies based on probe vehicle data~\cite{zhang2020,sekula2018} can reach comparable accuracy but rely on dense trajectory data that are not consistently available at national scale. Although Narayanan et al.~\cite{Makarov2024} reported higher $R^2$ values in a metropolitan case study, their analysis relied on synthetic traffic data rather than real-world observations. In contrast, DeepDemand is evaluated on 5,088 real highway segments across a national network, spanning multiple road classes and traffic regimes, making the prediction task both larger in scale and more heterogeneous.

Beyond predictive accuracy, a central contribution of DeepDemand lies in its interpretability. The learned travel-time deterrence function exhibits a clear and intuitive decay pattern, with a sharp reduction in car-use likelihood between approximately 15 and 45 minutes and a long but low-probability tail for longer trips. This structure is stable across cross-validation folds, while small regional shifts under spatial cross-validation suggest modest geographic variation in travel-time sensitivity. These patterns align well with established formulations of travel-time deterrence and empirical evidence on car-based travel behaviour, and demonstrate that the model learns meaningful behavioural structure without imposing predefined functional forms~\cite{4step}. Importantly, the pronounced decline in interaction strength beyond roughly 15 minutes of driving time provides quantitative support for time-based planning paradigms such as the “X-minute city”, where accessibility within short travel times is prioritised. Unlike traditional deterrence functions derived from travel surveys or imposed parametrically, the relationship observed here emerges directly from large-scale traffic data through theory-informed learning. 

The SHAP analysis further shows that OD pair scores are systematically associated with population structure, land use, employment composition, accessibility, and socioeconomic conditions. Higher car ownership, residential land use, and transport-related activity are associated with increased OD scores, while higher population density and younger age groups are associated with lower scores. The negative association with total population is consistent with gravity-based models such as DeepGravity~\cite{DeepGravity}, reflecting the tendency for highly populated urban areas to contribute proportionally less to highway traffic than lower-density, peripheral areas. Age-related patterns align with known differences in car ownership, travel autonomy, and time availability across population groups, particularly for highway-based trips~\cite{NTS2024}. More broadly, the observed relationships between car availability, deprivation, and OD scores are consistent with established evidence from UK travel surveys and the wider transport literature~\cite{EWfactorsaffectingtraffic,Goodwin2004,Lucas2012}. Importantly, DeepDemand recovers these well-documented associations through a fully data-driven OD representation, rather than through link-level regressions or manually specified behavioural rules, allowing asymmetric origin–destination effects to be examined explicitly.

Aggregating OD scores to origin and destination potentials reveals clear and interpretable spatial structures in trip generation and attraction. High destination-potential clusters align with major employment centres, transport hubs, and inter-urban corridors, forming a polycentric national pattern. City-level analyses further indicate that DeepDemand distinguishes destinations based on their functional relevance to the highway network rather than their nominal size or prominence. These results demonstrate that the model is able to infer latent interactions between land use, infrastructure provision, and travel behaviour, despite being trained only on aggregated edge-level traffic volumes and without direct supervision at the OD or area level.

Because DeepDemand relies exclusively on external socioeconomic inputs, it can be naturally extended to forward-looking scenario analysis. Using official population projections, we show that future traffic growth is likely to concentrate on motorways and major corridors connecting large metropolitan areas, while many rural or peripheral roads experience limited change or decline (Supplementary Appendix S10). This illustrates how the model can be used to explore spatially differentiated demand trajectories under alternative demographic scenarios, supporting strategic assessments of where future pressure on the highway network is most likely to emerge.

DeepDemand has several practical applications. The empirically learned travel-time sensitivity provides a behavioural benchmark for accessibility planning, revealing the critical time window within which most interaction intensity is concentrated. This offers a quantitative foundation for evaluating spatial equity in access to employment and services, and for assessing whether proposed infrastructure investments meaningfully expand functional urban reach. DeepDemand can support long-term transport planning by identifying corridors and regions where traffic demand is expected to intensify under projected demographic change. It can also be used for comparative scenario analysis, assessing how alternative population distributions or network configurations may redistribute demand across the highway system. The identification of high-potential origin and destination areas provides a data-driven basis for prioritising investment, congestion mitigation, and network resilience planning. Finally, because DeepDemand does not depend on historical traffic time series or probe data, it is well suited for deployment in data-sparse contexts, including regions with limited monitoring infrastructure or for early-stage planning exercises where traffic observations are unavailable.

Despite its strengths, several limitations should be noted. First, modal split is not represented explicitly, even though it is a component of classical TDM frameworks. While the model captures some modal substitution effects implicitly through socioeconomic and accessibility variables, it cannot directly evaluate policies that target mode choice, particularly in dense urban areas. Second, trip assignment is approximated using a static, learned function based on shortest paths rather than a dynamic equilibrium formulation, which may oversimplify route choice under congestion. Third, uncertainty in predictions and short-term temporal variability are not explicitly modelled. These limitations do not undermine the main findings but indicate directions for further development.

Future work could extend DeepDemand by incorporating explicit modal split representations, equilibrium-based or dynamic assignment mechanisms, and finer temporal resolution to capture peak-period dynamics. Integrating uncertainty quantification would further enhance its value for decision-making. In addition, cross-country transfer learning and few-shot calibration could be explored to assess generalisability beyond the UK context. Together, these extensions would enhance the model’s behavioural richness and policy relevance while preserving its core strengths of scalability and interpretability.

\section{Conclusion}\label{sec:conclusion}

This study introduces DeepDemand, a theory-informed deep learning framework for modelling highway traffic volumes using socioeconomic context and road-network structure. The framework combines a competitive two-source Dijkstra procedure for local origin–destination (OD) region extraction and screening with a differentiable architecture that models OD interactions and travel-time deterrence. This design embeds key components of classical travel demand models while allowing the relationships between origins, destinations, and travel cost to be learned directly from observed traffic data. As a result, the approach integrates behavioural structure and data-driven learning within a unified modelling framework.

Empirical evaluation on the UK strategic road network demonstrates that DeepDemand achieves strong predictive performance across a heterogeneous national network. The model consistently outperforms conventional baselines. Performance remains stable under spatial cross-validation, indicating that the framework captures demand relationships that generalise across geographic regions.

Beyond predictive accuracy, the explicit structure of DeepDemand enables interpretable analysis of travel demand mechanisms. The learned travel-time deterrence function exhibits a stable nonlinear decay pattern, with interaction intensity declining sharply between approximately 15 and 45 minutes of driving time. Feature-attribution analyses further indicate that OD interaction strength is strongly associated with demographic composition, car availability, employment structure, and land-use characteristics. Aggregated OD scores reveal coherent spatial patterns of trip generation and attraction, highlighting major demand centres within the national highway system.

Overall, these results demonstrate that theory-informed deep learning can provide an interpretable and effective alternative to conventional travel demand modelling approaches for estimating link-level traffic volumes. By relying on widely available socioeconomic and network data, the DeepDemand framework can support large-scale traffic estimation, accessibility analysis, and scenario-based transport planning, particularly in contexts where detailed travel surveys or historical traffic time series are limited.

\section*{Acknowledgements}
This research was supported by the Cambridge Commonwealth, European and International Trust. Additional support was provided by the Martin Centre for Architectural and Urban Studies and Cambridge Public Health. The authors gratefully acknowledge the Office for National Statistics, OpenStreetMap contributors, and National Highways for providing the data used in this study. Computational resources were provided by the Cambridge Service for Data Driven Discovery (CSD3), operated by the University of Cambridge Research Computing Service, with support from Dell EMC and Intel under EPSRC Tier-2 funding (EP/T022159/1) and DiRAC funding from the Science and Technology Facilities Council. The authors also thank colleagues and external experts for valuable discussions and feedback. The views expressed are those of the authors and do not necessarily reflect those of the supporting institutions.

\section*{Data availability}
All raw data used in this study are publicly available from their original providers. The UK road network, land-use polygons and points of interest were derived from OpenStreetMap data distributed via Geofabrik \cite{geofabrik}. Socioeconomic and demographic statistics at the LSOA level, including population, employment, household composition and car ownership, were obtained from the Office for National Statistics (ONS) \cite{population, employment, households, areapopden}. Income and employment deprivation indicators were obtained from the English Indices of Multiple Deprivation 2019 \cite{imd2019}. Ground-truth traffic volumes were derived from the National Highways Traffic Information System (TRIS) \cite{TRIS2024}. The processed datasets generated in this study, including the LSOA feature banks, node--LSOA mappings, screened OD pairs for each target edge, and AADT traffic volumes, are available in the University of Cambridge Apollo repository at \url{https://doi.org/10.17863/CAM.127892}.

\section*{Code availability}
All modelling code and data-processing scripts required to reproduce the results of this study are publicly available at \url{https://github.com/yueli901/DeepDemand}. The repository includes documentation describing the system configuration, core software dependencies, and the structure of the data and code directories. The code is released under the MIT License.

\bibliographystyle{elsarticle-num} 
\bibliography{reference.bib}

\end{document}